\documentclass[10pt,twocolumn,letterpaper]{article}

\usepackage{wacv}
\usepackage{times}
\usepackage{epsfig}
\usepackage{graphicx}
\usepackage{amsmath}
\usepackage{amssymb}
\usepackage{booktabs}
\usepackage{algorithm}
\usepackage{color} 
\usepackage{pythonhighlight}
\usepackage{wrapfig}
\usepackage{lipsum}
\usepackage{graphicx}
\usepackage{multirow}
 \RequirePackage{makecell}
\definecolor{forestgreen}{RGB}{34,139,34}

\def\wacvPaperID{8} 

\wacvalgorithmstrack   

\wacvfinalcopy 

\ifwacvfinal
\usepackage[breaklinks=true,bookmarks=false]{hyperref}
\else
\usepackage[pagebackref=true,breaklinks=true,colorlinks,bookmarks=false]{hyperref}
\fi

\begin{document}

\title{\textit{MixGen}: A New Multi-Modal Data Augmentation}


\author{%
  Xiaoshuai Hao\thanks{Work done during an internship at AWS. First three authors contribute equally.}\\
  Institute of Information Engineering, CAS\\
  \and
  Yi Zhu\\
  Amazon Web Services \\
  \and
  Srikar Appalaraju\\
  Amazon Web Services \\
  \and
  Aston Zhang \\
  Amazon Web Services \\
  \and
  Wanqian Zhang, Bo Li\\
  Institute of Information Engineering, CAS \\
  \and
  Mu Li \\
  Amazon Web Services
}

\maketitle
\thispagestyle{empty}

\begin{abstract}
   Data augmentation is a necessity to enhance data efficiency in deep learning. For vision-language pre-training, data is only augmented either for images or for text in previous works. In this paper, we present MixGen: a joint data augmentation for vision-language representation learning to further improve data efficiency. It generates new image-text pairs with semantic relationships preserved by interpolating images and concatenating text. It's simple, and can be plug-and-played into existing pipelines. We evaluate MixGen on four architectures, including CLIP, ViLT, ALBEF and TCL, across five downstream vision-language tasks to show its versatility and effectiveness. For example, adding MixGen in ALBEF pre-training leads to absolute performance improvements on downstream tasks: image-text retrieval (+$6.2\%$ on COCO fine-tuned and +$5.3\%$ on Flicker30K zero-shot), visual grounding (+$0.9\%$ on RefCOCO+), visual reasoning (+$0.9\%$ on NLVR$^{2}$), visual question answering (+$0.3\%$ on VQA2.0), and visual entailment (+$0.4\%$ on SNLI-VE).
\end{abstract}

\section{Introduction}
\label{sec:introduction}

Recent years have witnessed an explosion in vision-language research ~\cite{vilbert_Lu2019ViLBERTPT,2020_iclr_vlbert,2020_eccv_uniter,Li_2022_WACV,2021_arxiv_imagebert,2021_cvpr_sob,2021_arxiv_vlmo,Appalaraju_2021_ICCV}. 
In joint modality learning, models extract rich information across  modalities to learn better latent representations.
These models, however, are often trained on a massive number of image-text pairs using thousands of GPUs.
For example,  CLIP~\cite{2021_arxiv_clip} matches the accuracy of ResNet-50~\cite{2016_cvpr_resnet} on ImageNet by only using zero-shot, but it is trained with 400M image-text pairs for 12 days on 256 V100 GPUs. 
Furthermore, most of these large-scale datasets~\cite{2021_arxiv_clip,2021_icml_align,2021_arxiv_florence} are not publicly accessible. Even if they are available, reproduction and further improvement on existing methods are challenging for researchers with limited computing resources. 

Data augmentation is widely used in deep learning to improve data efficiency and provide explicit regularization during model training in both computer vision (CV)~\cite{2017_arxiv_cutout,2018_iclr_mixup,2019_cvpr_autoaugment,2019_iccv_cutmix,2020_neurips_randaugment,2020_aaai_randomerasing,tang_iccv2021_sncn} and natural language processing (NLP)~\cite{zhang2015character,wang2015s,jiao2019tinybert,xie2020unsupervised, wei2019eda,coulombe2018text,regina2020text,yan2019data}. Applying existing data augmentation techniques to vision-language learning, however, is not straightforward. In an image-text pair, both the image and the text contain rich information that matches each other. Intuitively, we hope that their semantics still match after data augmentation. For example, consider an image with its paired sentence ``a \textit{white} dog playing in the \textit{right} corner of the \textit{green} lawn''. 
Applying data augmentation methods such as
cropping, color changing, and flipping to this image may require changing the color and positional words in its paired sentence at the same time.

In order to preserve the semantic relationship, previous works perform mild data augmentation on \emph{either} vision \emph{or} text modality. 
ViLT~\cite{2021_icml_vilt} and following  works~\cite{2021_arxiv_vlmo,2021_neurips_albef} adopt RandAugment~\cite{2020_neurips_randaugment} for image augmentations without color inversion.
CLIP~\cite{2021_arxiv_clip} and ALIGN~\cite{2021_icml_align} only use random resized crop without other image augmentations.
In terms of language, most literature just leaves text data augmentation to be handled by masked language modeling~\cite{2019_naacl_bert}.
There are works using co-augmentation~\cite{2020_eccv_seadavqa,2021_intelligent_mda}, but only designed for specific downstream tasks, rather than generic vision-language pre-training.

\begin{figure*}
\begin{minipage}{0.4\textwidth}
\centering
\includegraphics[width=1.0\linewidth]{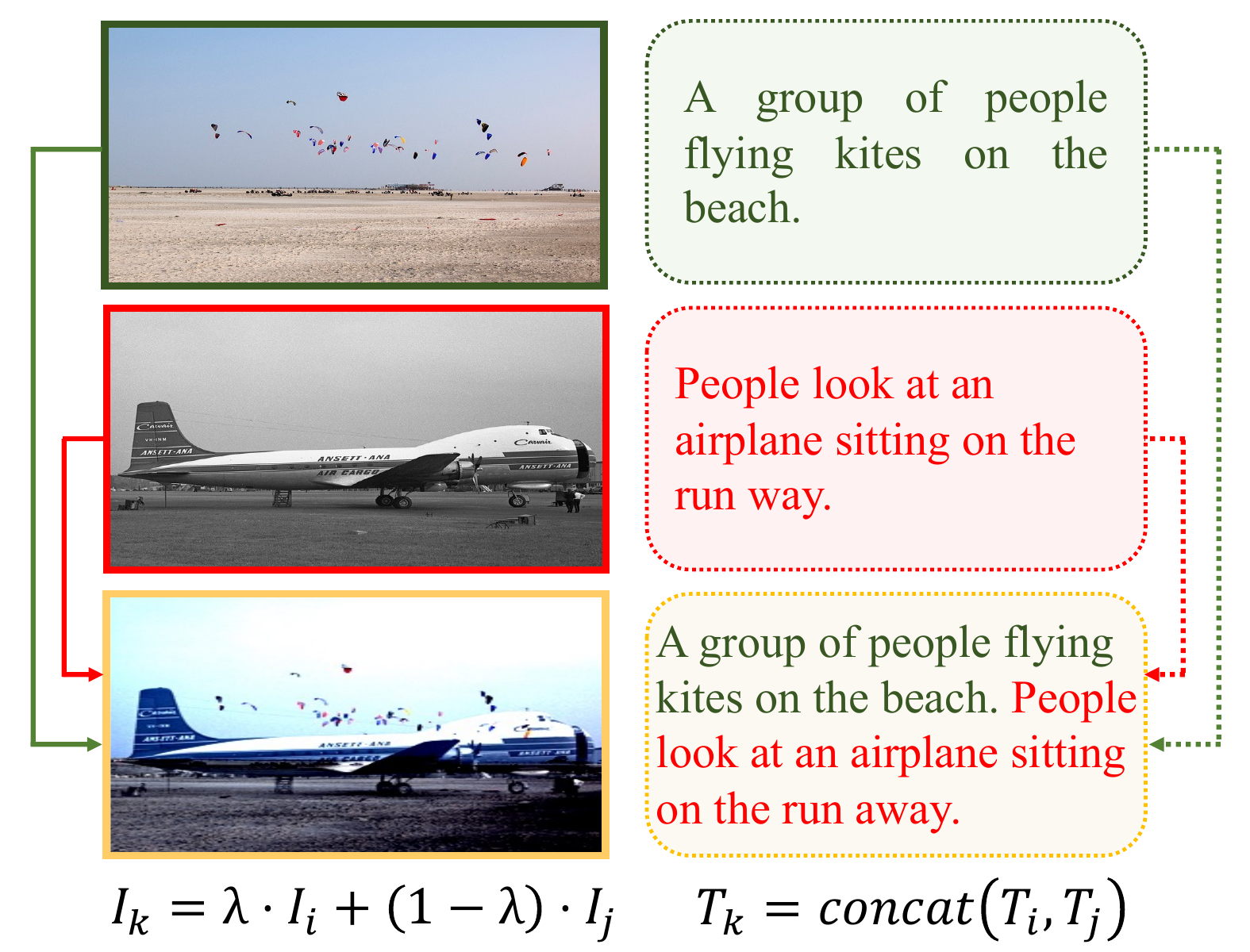}
\label{fig:mixgen}
\end{minipage}
\hfill
\begin{minipage}{0.57\textwidth}
\vspace{-4ex}
\begin{algorithm}[H]
\scriptsize
\caption{Pseudocode of MixGen data augmentation}

\begin{ttfamily}

\textcolor{teal}{\# image, text: a batch of B randomly sampled $\{$image, text$\}$}

\textcolor{teal}{\# M: number of new image-text pairs generated by MixGen}

\textcolor{teal}{\# $\lambda$: image mixup ratio}

\texttt{\\}
def mixgen(image, text, M, $\lambda$):

\quad \,\, for i in range(M):

\qquad \quad \textcolor{teal}{\# image mixup}

\qquad \quad image[i,:] = ($\lambda$ * image[i,:] + (1-$\lambda$) * image[i+M,:]) 

\qquad \quad  \textcolor{teal}{\# text concatenation}

\qquad \quad  text[i] = text[i] + " " + text[i+M]  \quad

\quad \,\,  return image, text
\end{ttfamily}
\end{algorithm}
\end{minipage}
\caption{Our proposed multi-modal data augmentation technique: \textbf{MixGen}. Given any two image-text pairs $(I_{i}, T_{i})$ and $(I_{j}, T_{j})$, we interpolate the two images and concatenate two text sequences, to generate a new image-text pair $(I_{k}, T_{k})$. Figure best viewed in color.}
\end{figure*}

In this work, we introduce a multi-modal joint data augmentation method for pre-training: \textbf{Mix} \textbf{Gen}eration (MixGen).
As shown in Figure~\ref{fig:mixgen}, MixGen generates a new training sample by linearly interpolating images and concatenating text sequences from two existing image-text pairs.
We can see that most objects and scene layout remain in the blended image, while the text information is fully preserved. 
The semantic relationship within the newly generated image-text pair is expected to match in most cases.
Thus, we can use the augmented data to improve vision-language model training. 

Despite its simplicity, using MixGen on top of strong baselines (e.g., ALBEF~\cite{2021_neurips_albef}) consistently improves state-of-the-art performance across five downstream vision-language tasks: image-text retrieval (+$6.2\%$ on COCO fine-tuned and +$5.3\%$ on Flicker30K zero-shot), visual grounding (+$0.9\%$ on RefCOCO+), visual reasoning (+$0.9\%$ on NLVR$^{2}$), visual question answering (+$0.3\%$ on VQA2.0), and visual entailment (+$0.4\%$ on SNLI-VE).
MixGen also leads to enhanced data efficiency, e.g., the performance of ALBEF with MixGen when pre-trained on 1M/2M/3M samples match baseline ALBEF pre-trained on 2M/3M/4M samples, respectively. 
In addition, we perform extensive ablation studies to understand the effects of various design choices in MixGen.
Finally, MixGen can be incorporated into most methods with only a few lines of code. 
In terms of fine-tuned image-text retrieval on COCO, MixGen brings absolute improvements on four popular and varied architectures: ViLT (+$17.2\%$), CLIP (+$4.1\%$), ALBEF (+$7.0\%$) and TCL (+$3.2\%$).

\section{MixGen}
\label{sec:mixgen}

\begin{figure*}[t]
\centering
	\includegraphics[width=1.0 \linewidth]{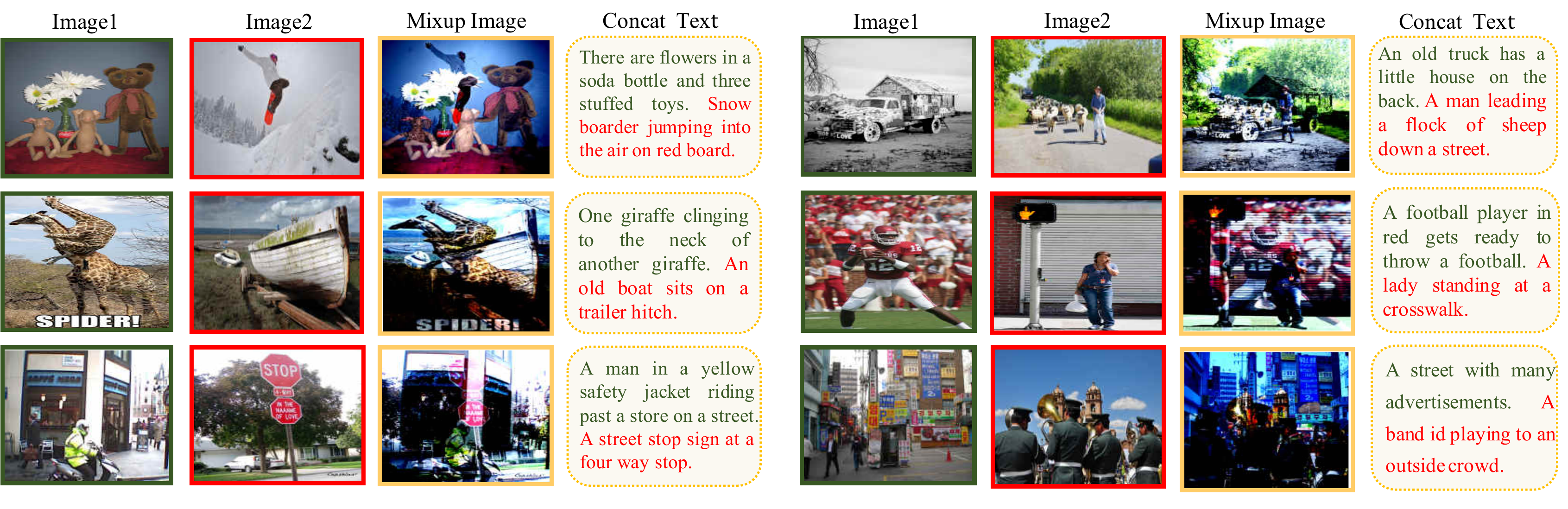}
	\caption{\textbf{More image-text pairs generated by MixGen}. The new pairs not only preserve original semantic relationships, but also increase diversity to model training. Figure best viewed in color.
	\label{fig:more_mixgen_samples}}
\end{figure*}

In this section, we introduce our multi-modal joint data augmentation technique: \textbf{Mix} \textbf{Gen}eration (MixGen).
Suppose that we have a dataset of $N$ image-text pairs, where images and text are denoted as $I$ and $T$ with subscripts, respectively.
Given two image-text pairs $(I_{i}, T_{i})$ and $(I_{j}, T_{j})$ for any $i,j \in \{1, \ldots, N\}$ and $i \neq j$, a new training sample $(I_{k}, T_{k})$ is generated via
\begin{equation}
\begin{split}
    I_{k} &= \lambda \cdot I_{i} + (1 - \lambda) \cdot I_{j} \\ 
    T_{k} &= \mathrm{concat}(T_{i}, T_{j}),
\end{split}
\label{equ:input_mixgen}
\end{equation}
where $\lambda$ is a hyper-parameter between 0 and 1, indicating linear interpolation between raw pixels of two images $I_{i}$ and $I_{j}$;
and the $\mathrm{concat}$ operator directly concatenates two text sequences $T_{i}$ and $T_{j}$ to best preserve original information.
In this way, the semantic relationship within the newly generated image-text pair $(I_{k}, T_{k})$ still holds in most scenarios, such as in Figure~\ref{fig:mixgen} and Figure~\ref{fig:more_mixgen_samples}. This random combination of image-text samples also increase diversity to model training, which elicits making available rare concepts.

The pseudocode of MixGen data augmentation is presented in Algorithm 1.
Given a minibatch of $B$ randomly sampled image-text pairs, MixGen replaces the first $M$ training samples with the newly generated pairs.
Hence, the batch size, total training iterations and overall training pipeline remain the same.
By default, we set $\lambda=0.5$ and $M=B/4$ in Algorithm 1.
Such a plug-and-play technique can be easily incorporated into most vision-language representation learning approaches and tasks: all it takes is just a few lines of code with minimal computational overhead.
Take ALBEF~\cite{2021_neurips_albef} as an example, adding MixGen on top of it during pre-training only increases its overall wall clock training time by $0.4\%$. 

\subsection{MixGen variants}
\label{subsec:variants}

MixGen is in a very simple form (Algorithm 1). 
However, there could be multiple variants depending on how image and text augmentations are performed. 
Theoretically, we could also use other augmentations for images beyond mixup, and other text augmentations beyond concatenation, but the design space will be intractable.
Thus, we focus on using mixup for image and concatenation for text, and select 5 most straightforward variants of MixGen to support our final design choice.

Since default MixGen takes a fixed $\lambda$, we introduce variant (a) with $\lambda \sim$ Beta(0.1, 0.1), following original mixup~\cite{2018_iclr_mixup} that samples $\lambda$ from a Beta distribution.
In order to show the benefit of performing joint image-text augmentation, we propose variant (b) and (c).  
To be specific, variant (b) mixup two images and uniformly pick one text sequence, while variant (c) concatenates two text sequences and uniformly pick one image. 
In the end, we study whether we should use a subset of tokens, instead of concatenating all tokens from two text sequences.
Variant (d) takes tokens from two text sequences proportionally based on $\lambda$ similar as image mixup and then concatenate. 
The other variant (e) first concatenates all the tokens, but randomly keep half of them to generate a new text sequence.

More detailed definitions of these 5 variants can be seen in Table~\ref{tab:mixgen_variants}.
We also perform extensive ablation studies on them.
As we will see in Table~\ref{table:variants}, our default MixGen achieves the overall best performance, and consistently outperforms other variants across four different vision-language downstream tasks.

\subsection{Input-level and embedding-level MixGen}
\label{subsec:input_embedding}

Another design perspective is where to apply the data augmentation. 
The formulation in Equation~(\ref{equ:input_mixgen}) is directly performed on the raw input, e.g., images and text sequences.
Alternatively, the idea of MixGen can be applied on the embedding level.
To be specific, instead of interpolating raw image pixels, we can interpolate image features that are extracted from an image encoder. 
Similarly, instead of concatenating two text sequences, we can concatenate two sequence features that are extracted from a text encoder. 
Denoting the training pairs with respect to their embedding as $(f_{I_{i}}, f_{T_{i}})$ and $(f_{I_{j}}, f_{T_{j}})$, the newly generated training pair in its embedding form is
\begin{equation}
\begin{split}
    f_{I_{k}} &= \lambda \cdot f_{I_{i}} + (1 - \lambda) \cdot f_{I_{j}} \\ 
    f_{T_{k}} &= \text{concat}(f_{T_{i}}, f_{T_{j}}).
\end{split}
\label{equ:embedding_mixgen}
\end{equation}

We call MixGen that is performed on the raw input as \textit{input-level MixGen}, and that performed on the embedding level as \textit{embedding-level MixGen}. 
As we will show in Figure~\ref{fig:optimal_embedding} left, input-level MixGen consistently performs better than embedding-level MixGen. 
In addition, input-level MixGen has an advantage of implementational simplicity, since we don't need to modify network architecture nor touch model forward.
Hence, we refer to \textit{input-level MixGen} as our proposed MixGen for the rest of the paper unless otherwise stated.

\begin{table*}[t]
\begin{center}
\caption{\textbf{Fine-tuned image-text retrieval} on Flickr30K and MSCOCO datasets.\label{tab:ft_retrieval}}
\vspace{1ex}
\resizebox{\textwidth}{!}{
  \begin{tabular}{l|c|ccccccl|ccccccl}
  \multirow{3}{*}{Method}&
  \multirow{3}{*}{$\#$Images}&
  \multicolumn{7}{c|}{MSCOCO(5K test set)} & \multicolumn{7}{c}{Flickr30K(1K test set)} \\

   & &\multicolumn{3}{c}{Text Retrieval} & \multicolumn{3}{c}{Image Retrieval} & \multicolumn{1}{c|}{} & \multicolumn{3}{c}{Text Retrieval} & \multicolumn{3}{c}{Image Retrieval}&\multicolumn{1}{c}{} \\
   &  &R@1  & R@5 & R@10& R@1  & R@5 & R@10 & RSUM & R@1  & R@5 & R@10& R@1  & R@5 & R@10 & RSUM\\
  \midrule
  UNITER-base~\cite{2020_eccv_uniter}&4M&64.4&87.4&93.1&50.3&78.5&87.2&460.9&85.9&97.1&98.8&72.5&92.4&96.1&542.8\\
  VILLA-base~\cite{2020_neurips_villa}&4M& - & - & - & - & -  & - & - &86.6&97.9&99.2&74.7&92.9&95.8&547.1\\
  OSCAR-base~\cite{2020_eccv_oscar}&4M&70.0&91.1&95.5&54.0&80.8&88.5&479.9& - & - & - & - & - & - & - \\
  UNIMO-base~\cite{2021_acl_unimo}&4M& - & - & - & - & - & - & - &89.7&98.4&99.1&74.7&93.4&96.1&551.4\\
  ViLT-base~\cite{2021_icml_vilt}&4M&61.5&86.3&92.7&42.7&72.9&83.1&439.2&83.5&96.7&98.6&64.4&88.7&93.8&525.7\\
  \color{gray}ALBEF-base~\cite{2021_neurips_albef}&\color{gray}4M&\color{gray}73.1&\color{gray}91.4&\color{gray}96.0&\color{gray}56.8&\color{gray}81.5&\color{gray}89.2&\color{gray}488.0&\color{gray}94.3&\color{gray}99.4&\color{gray}99.8&\color{gray}82.8&\color{gray}96.7&\color{gray}98.4&\color{gray}571.4\\
    \midrule
  ALBEF-base~\cite{2021_neurips_albef}& 3M &72.5&91.7&95.9&55.8&81.3&88.4&485.6& $\textbf{95.1}$ &99.1&99.7&81.4&96.0& $\textbf{98.2}$ &569.5\\
  ALBEF-base+$\textbf{MixGen}$ & 3M & $\textbf{74.2}$ & $\textbf{92.8}$ & $\textbf{96.4}$ & $\textbf{57.3}$ & $\textbf{82.1}$ & $\textbf{89.0}$ & $\textbf{491.8}_{\textcolor{forestgreen}{+6.2}}$ & 94.8 & $\textbf{99.4}$ & $\textbf{100.0}$ & $\textbf{82.4}$ & $\textbf{96.3}$ & 98.0 & $\textbf{570.9}_{\textcolor{forestgreen}{+1.4}}$\\
  \end{tabular}}
\end{center}
\vspace{-1ex}
\end{table*}

\begin{table}[t]
\begin{center}
\caption{\textbf{Zero-shot image-text retrieval} on Flickr30K dataset.\label{tab:zs_retrieval}}
\resizebox{0.5\textwidth}{!}{
\begin{tabular}{l|c|ccccccl}
\multirow{3}{*}{Method}&
\multirow{3}{*}{$\#$Images}&
\multicolumn{7}{c}{Flickr30K(1K test set)} \\

 &  & \multicolumn{3}{c}{Text Retrieval} & \multicolumn{3}{c}{Image Retrieval} & \multicolumn{1}{c}{} \\
 &  &R@1  & R@5 & R@10& R@1  & R@5 & R@10 & RSUM \\
\midrule
UNITER-base~\cite{2020_eccv_uniter}&4M&80.7&95.7&98.0&66.2&88.4&92.9&521.9\\
ViLT-base~\cite{2021_icml_vilt}&4M&73.2&93.6&96.5&55.0&82.5&89.8&490.3\\
\color{gray}ALBEF-base~\cite{2021_neurips_albef}&\color{gray}4M&\color{gray}90.5&\color{gray}98.8&\color{gray}99.7&\color{gray}76.8&\color{gray}93.7&\color{gray}96.7&\color{gray}556.2\\
\midrule
ALBEF-base~\cite{2021_neurips_albef}&3M&91.1&98.2&99.3&75.7&92.5&96.0&552.8\\
ALBEF-base+$\textbf{MixGen}$ & 3M & $\textbf{91.6}$ & $\textbf{99.2}$ & $\textbf{99.9}$ & $\textbf{77.2}$ & $\textbf{93.6}$ & $\textbf{96.6}$ & $\textbf{558.1}_{\textcolor{forestgreen}{+5.3}}$ \\
\end{tabular}}
\end{center}
\vspace{-2ex}
\end{table}

\section{Experiments}
\label{sec:experiments}

In this section, we first describe model pre-training in Sec.~\ref{subsec:pretrain}.
Then we introduce five downstream tasks (image-text retrieval, visual question answering, visual entailment, visual reasoning and visual grounding) and present experimental results of using MixGen in Sec.~\ref{subsec:retrieval} and Sec.~\ref{subsec:downstream}.
In the end, we show visualizations in Sec.~\ref{subsec:visualization} to demonstrate the benefits of using MixGen.

\subsection{Pre-training}
\label{subsec:pretrain}

\noindent \textbf{Pre-training methods}
Data augmentation techniques are usually applicable to various models, because they are agnostic to network architectures, training objectives, optimizers, or learning schedules.
In this study, we plug-and-play MixGen on four recent popular approaches with varied architectures, CLIP~\cite{2021_arxiv_clip} (dual-encoder), ViLT~\cite{2021_icml_vilt} (single fusion encoder),  ALBEF~\cite{2021_neurips_albef} (dual-encoder followed by fusion encoder) and TCL~\cite{TCL_Yang2022VisionLanguagePW} (dual-encoder followed by fusion encoder). Also note, the pre-training objectives for these approaches are varied. 
We emphasize that MixGen can be easily incorporated to other vision-language pre-training methods with only a few lines of code. 
To be specific, once we get a minibatch out of the dataloader, we apply MixGen to generate  new image-text pairs and update the minibatch. We then forward the updated batch to the network without modifying any of their original training settings.

\noindent \textbf{Pre-training datasets}
There are four widely adopted datasets for vision-language model pre-training, including COCO~\cite{2014eccvcoco}, Visual Genome (VG)~\cite{2017vgdataset}, SBU Captions~\cite{2011_neurips_sbu} and Conceptual Captions (CC)~\cite{2021cvprgcc12m}. 
Most literature~\cite{2020_eccv_uniter,2021_arxiv_vlmo,2021_icml_vilt,2021_neurips_albef} used a combination of the four datasets as the standard pre-training setting, which leads to a total of 4M unique images and 5.1M image-text pairs. 
However, for SBU and CC datasets, image-text pairs are provided in url format, and a portion of the urls are unaccessible. 
For example, we can only download 2.2M out of 3M images of CC dataset. 
Hence, our final training set only has 3.3M unique images and 4.4M image-text pairs.
Compared with previous literature that has access to the original 5.1M image-text pairs, we have about 700K less training samples.
In this paper, we term the original setting as 4M, and ours as 3M. 

\noindent \textbf{Pre-training implementation details}
Majority of our experiments are performed based on ALBEF~\cite{2021_neurips_albef} given its superior performance on a range of downstream tasks. 
We adopt the official codebase\footnote{https://github.com/salesforce/ALBEF} kindly provided by the authors.
Following~\cite{2021_neurips_albef}, the model is trained for $30$ epochs with a total batch size of $512$ on 8 V100 GPUs. 
When using MixGen, the default value of $M$ is set to a quarter of the batch size $B$, i.e., a minibatch will contain $384$ existing samples and $128$ new image-text pairs.
AdamW~\cite{2019_iclr_optimizer} is adopted as the optimizer with a weight decay of $0.02$.
Learning rate is warmed-up to $1e^{-4}$ in the first $1000$ iterations and then decayed to $1e^{-5}$ following a cosine schedule.
For ViLT and TCL, we use their official codebase\footnote{https://github.com/dandelin/ViLT} \footnote{https://github.com/uta-smile/TCL}.
For CLIP, we adopt a reproducible implementation from open-clip\footnote{https://github.com/mlfoundations/open$\_$clip}. 
To save memory and speed up experiments, we use mix precision training~\cite{2018_iclr_fp16}.

\begin{table}[t]
\begin{center}
\caption{\textbf{Compatibility to other vision-language pre-training methods}. Adding MixGen leads to consistent performance boost in terms of image-text retrieval. Note the models here are pre-trained on three datasets (COCO, VG and SBU) with 1M images. Ft: fine-tuned setting. Zs: zero-shot setting.\label{table:general_model}
\vspace{2ex}}
\scalebox{0.6}{
  \begin{tabular}{l|c|l|l|l}
  \toprule
  \multirow{2}{*}{Methods}&\multirow{2}{*}{Venue}&\makecell[c]{Flickr30K-Ft}&\makecell[c]{MSCOCO-Ft}&\makecell[c]{Flickr30K-Zs} \\
    & &\makecell[c]{(1K test set)}&\makecell[c]{(5K test set)}&\makecell[c]{(1K test set)}\\
     \midrule
  ViLT~\cite{2021_icml_vilt}&ICML 21&232.2&172.5&135.5\\
  ViLT + $\textbf{MixGen}$ &-& $\textbf{241.4}_{\textcolor{forestgreen}{+9.20}}$ & $\textbf{189.7}_{\textcolor{forestgreen}{+17.2}}$ & $\textbf{150.4}_{\textcolor{forestgreen}{+14.9}}$ \\
  CLIP~\cite{2021_arxiv_clip}&ICML 21&538.2&450.5&485.3\\
  CLIP + $\textbf{MixGen}$ &-& $\textbf{541.0}_{\textcolor{forestgreen}{+2.80}}$ & $\textbf{454.6}_{\textcolor{forestgreen}{+4.10}}$ & $\textbf{492.1}_{\textcolor{forestgreen}{+6.80}}$ \\
  ALBEF~\cite{2021_neurips_albef}&NeurIPS 21&555.7&470.7&518.0\\
  ALBEF + $\textbf{MixGen}$ &-& $\textbf{561.0}_{\textcolor{forestgreen}{+5.30}}$ & $\textbf{477.7}_{\textcolor{forestgreen}{+7.00}}$ & $\textbf{524.3}_{\textcolor{forestgreen}{+6.30}}$ \\
  TCL~\cite{TCL_Yang2022VisionLanguagePW}&CVPR 22&561.5&487.0&537.3\\
  TCL + $\textbf{MixGen}$ &-& $\textbf{563.6}_{\textcolor{forestgreen}{+2.10}}$ & $\textbf{490.2}_{\textcolor{forestgreen}{+3.20}}$ & $\textbf{539.5}_{\textcolor{forestgreen}{+2.20}}$ \\
  \bottomrule
  \end{tabular}}
\end{center}
\vspace{-2ex}
\end{table}

\subsection{Image-text retrieval}
\label{subsec:retrieval}
Image-text retrieval includes two subtasks: (1) retrieve images with given text (Image Retrieval) and (2) retrieve text with given images (Text Retrieval).
We conduct experiments on MSCOCO~\cite{2014eccvcoco} and Flickr30K~\cite{2015iccvflick30} datasets.
In terms of evaluation, we use recall at K $(R@K)$ metric, where $K = \{1, 5, 10\}$.
For the purpose of easy comparison, we also use RSUM as the metric to reveal the overall performance of models following~\cite{Wu_2019_CVPR}, which is defined as the sum of recall metrics at $K = \{1, 5, 10\}$ of both image and text retrieval tasks.

\noindent \textbf{Fine-tuned results} In Table~\ref{tab:ft_retrieval}, we can see that MixGen consistently improves over our ALBEF baseline on both datasets.
Under the 3M setting, simply adding MixGen without any modifications leads to an improvement of $6.2\%$ RSUM score on COCO and $1.4\%$ RSUM score on Flicker30K. 
Note that, due to the missing data problem, our reproduced ALBEF (trained on 3.3M pairs) achieves marginally lower performance than the numbers reported in the original paper (trained on 4M pairs, {\color{gray}gray} row in Table~\ref{tab:ft_retrieval}).
However, after adding MixGen, our performance is even better than the original paper on COCO and competitive on Flicker30K, despite our model is trained with 700K less image-text pairs.
This clearly shows that MixGen improves the data efficiency for model training.

\begin{table*}[t]
\begin{center}
\caption{\textbf{Comparison with state-of-the-art methods on downstream vision-language tasks}. MixGen consistently improve across VQA, VR and VE. }
\label{tab:downstream}
\scalebox{0.8}{
\begin{tabular}{l|c|ll|ll|ll}   
\multirow{2}{*}{Method}&\multirow{2}{*}{$\#$Images}&
\multicolumn{2}{c}{VQA} & \multicolumn{2}{c}{NLVR$^{2}$} & \multicolumn{2}{c}{SNLI-VE} \\
 &  & test-dev  & test-std & dev& test-P   & val& test  \\
\midrule
VisualBERT-base~\cite{2019_arxiv_visualbert}&4M&70.80&71.00&67.40&67.00&-&-\\
OSCAR-base~\cite{2020_eccv_oscar}&4M&73.16&73.44&78.07&78.36&-&-\\
UNITER-base~\cite{2020_eccv_uniter}&4M&72.70&72.91&77.18&77.85&78.59&78.28\\
ViLT-base~\cite{2021_icml_vilt}&4M&71.26&-&75.70&76.13&-&-\\
UNIMO-base~\cite{2021_acl_unimo}&4M&73.79&74.02&-&-&80.00&79.10\\
VILLA-base~\cite{2020_neurips_villa}&4M&73.59&73.67&78.39&79.30&79.47&79.03\\
\color{gray}ALBEF-base~\cite{2021_neurips_albef}&\color{gray}4M&\color{gray}74.54&\color{gray}74.70&\color{gray}80.24&\color{gray}80.50&\color{gray}80.14&\color{gray}80.30\\
\midrule
ALBEF-base~\cite{2021_neurips_albef}&3M&74.38&74.51&79.47&80.05&79.49&79.69\\
ALBEF-base+$\textbf{MixGen}$ &3M& $\textbf{74.51}_{\textcolor{forestgreen}{+0.13}}$ & $\textbf{74.79}_{\textcolor{forestgreen}{+0.28}}$ & $\textbf{80.23}_{\textcolor{forestgreen}{+0.76}}$ & $\textbf{80.94}_{\textcolor{forestgreen}{+0.89}}$ & $\textbf{80.05}_{\textcolor{forestgreen}{+0.56}}$ & $\textbf{80.05}_{\textcolor{forestgreen}{+0.36}}$ \\
\end{tabular}}
\end{center}
\end{table*}

\begin{figure*}[t]
\centering
\includegraphics[width=0.95\textwidth]{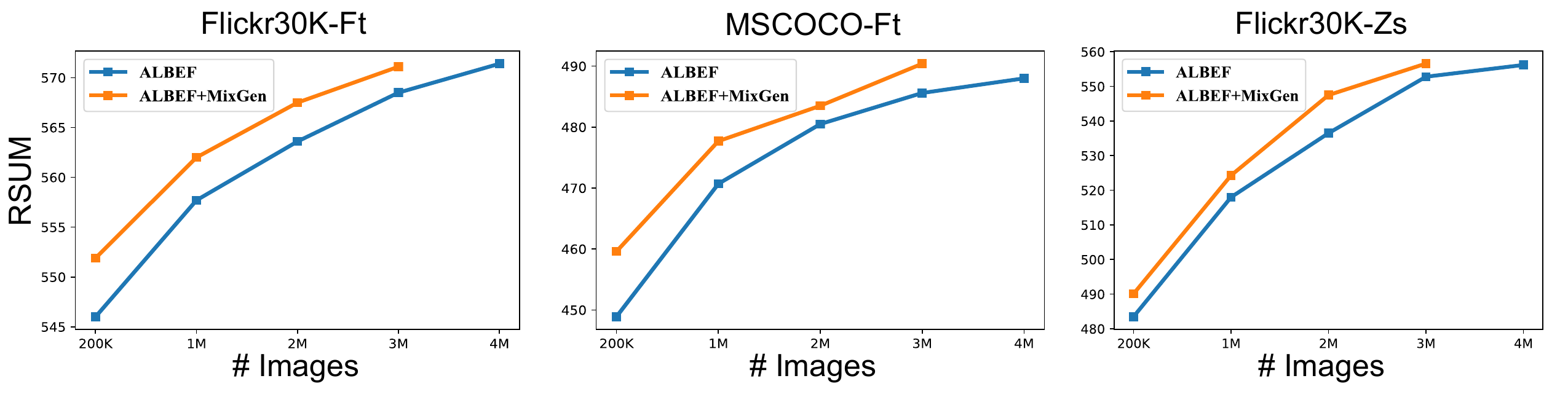}
\caption{\textbf{Image-text retrieval performance given various number of training images}. Note that ALBEF with MixGen trained on 3M images achieves competitive performance or outperform baseline ALBEF trained on 4M images. This indicates the data efficiency of MixGen as an effective data augmentation method.
}
\label{fig:image_number}
\end{figure*}

\noindent \textbf{Zero-shot results} In Table~\ref{tab:zs_retrieval}, similar conclusions can be observed. 
Under our 3M setting, MixGen leads to an improvement of $5.3\%$ RSUM score on Flicker30K. 
Since zero-shot setting treats pre-trained models as feature extractors, such significant performance gains suggest that the multi-modal features learned with MixGen during pre-training generalize well. 

\noindent \textbf{Compatibility with other VL models}
We show MixGen's compatibility with other vision-language pre-training methods, i.e., CLIP~\cite{2021_arxiv_clip}, ViLT~\cite{2021_icml_vilt} and TCL~\cite{TCL_Yang2022VisionLanguagePW}.
Besides adding MixGen, we do not modify their original training settings.
Given ViLT training is very costly (e.g., 3 days with 64 V100 GPUs), we only use three datasets (COCO, VG, and SBU) instead of four during pre-training for this experiment. The dataset consists of 1M unique images and 2.2M image-text pairs, which we term it the 1M setting.
As shown in Table~\ref{table:general_model}, simply adding MixGen on top of these strong baselines consistently improve state-of-the-art performance.
In terms of fine-tuned image-text retrieval on COCO, MixGen demonstrates significant accuracy boost (absolute): ViLT (+$17.2\%$), CLIP
(+$4.1\%$), ALBEF (+$7.0\%$) and TCL (+$3.2\%$).
This shows the versatility of MixGen as image-text data augmentation in pre-training.

\noindent \textbf{Data efficiency}
We investigate how much data efficiency MixGen can achieve. 
We reduce the number of unique images used for pre-training, from 3M to 2M, 1M and 200K.
We have described 1M and 3M settings. For 2M, we use three datasets plus a random subset from CC dataset. For 200K, we simply use two datasets (COCO and VG). 
The performance of image-text retrieval can be seen in Figure~\ref{fig:image_number}.
We first notice that adding MixGen is always better than without it. 
In particular, the improvement in low data regime is more significant.
Secondly, the performance of ALBEF with MixGen when trained on 1M, 2M and 3M samples match baseline ALBEF when trained on 2M, 3M and 4M samples, respectively. 
This again indicates the data efficiency of MixGen.

\begin{figure*}[t]
\centering
	\includegraphics[width=0.9 \linewidth]{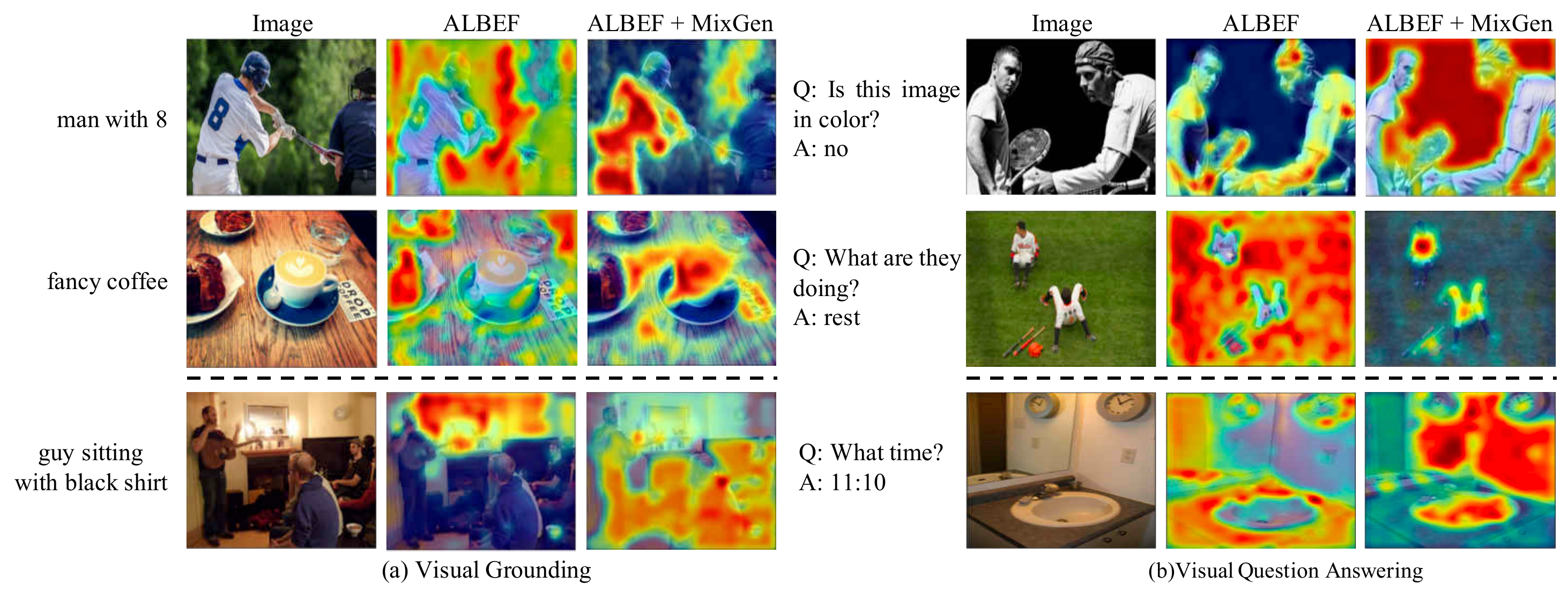}
	\caption{\textbf{Grad-CAM visualization for visual grounding and VQA}. First two rows are successful cases where MixGen helps, and the third row shows failure cases. Figure best viewed in color.\label{fig:vis_vg_vqa}}
\end{figure*}

\begin{figure}[t]
\centering
	\includegraphics[width=0.95 \linewidth]{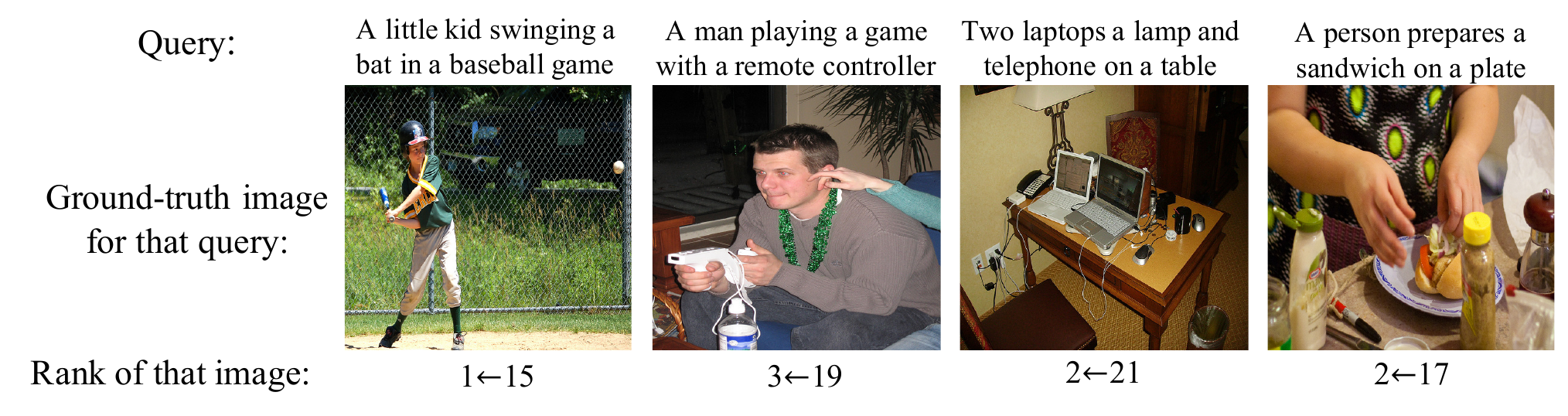}
	\caption{\textbf{Text-to-image retrieval on MSCOCO}. Given a text query, $A \leftarrow B$ indicates the rank of the retrieved ground-truth image improves from $B$ (wo MixGen) to $A$ (w MixGen).}
	\label{fig:vis_retrieval}
	\vspace{-1ex}
\end{figure}

\begin{table}
\begin{center}
\caption{\textbf{Weakly-supervised visual grounding} on RefCOCO+}
\vspace{1ex}
\label{tab:grounding}
\scalebox{0.8}{
\begin{tabular}{l|c|l|l|l}  
\makecell[l]{Method}&\makecell[c]{$\#$Images}&\makecell[l]{Val}&\makecell[l]{TestA}&\makecell[l]{TestB}\\
  \midrule
  ARN~\cite{2019_iccv_adaptive}&14M&32.78&34.35&32.13\\
  CCL~\cite{2022_NeruIPS_counterfactual}&14M&34.29&36.91&33.56\\
  \color{gray}ALBEF$_{itc}$~\cite{2021_neurips_albef}&\color{gray}14M&\color{gray}51.58&\color{gray}60.09&\color{gray}40.19\\
  \color{gray}ALBEF$_{itm}$~\cite{2021_neurips_albef}&\color{gray}14M&\color{gray}58.46&\color{gray}65.89&\color{gray}46.25\\
  \midrule
  ALBEF$_{itm}$~\cite{2021_neurips_albef}&3M&57.76&65.08&45.57\\
  ALBEF$_{itm}$+$\textbf{MixGen}$&3M&56.72& $\textbf{65.35}_{\textcolor{forestgreen}{+0.27}}$ & $\textbf{46.47}_{\textcolor{forestgreen}{+0.9}}$ \\
\end{tabular}}
\end{center}
\end{table}

\subsection{Evaluation on VQA, VR, VE and VG}
\label{subsec:downstream}
We first describe the other four downstream tasks. 
Visual Question Answering (VQA) aims to predict an answer given an image and a corresponding question.
We conduct experiments on the VQA 2.0 dataset~\cite{2017_cvpr_vqa2} and consider it as an answer generation task following~\cite{2021_neurips_albef,cho2021vlt5}.
We constrain the fine-tuned answer decoder to generate answers from the set of $3,192$ candidates for fair comparison to other literature. 
We use the official evaluation server\footnote{https://visualqa.org/evaluation.html} to report accuracy.
Visual Entailment (VE) is a visual reasoning task to predict whether the relationship between an image and text is entailment, neutral or contradictory. 
Following~\cite{2020_eccv_uniter,2021_neurips_albef}, we treat this task as a three-way classification problem and report classification accuracy on the SNLI-VE~\cite{2019vedataset} dataset.
Visual Reasoning (VR) requires the model to determine whether a textual statement describes a pair of images. 
We conduct experiments on the NLVR$^{2}$~\cite{2019NLVR2} dataset following~\cite{2021_icml_vilt,2021_neurips_albef,2020_eccv_oscar} and report standard per-example prediction accuracy.
Visual Grounding (VG) localizes an image region that corresponds to a text description. Here, we follow~\cite{2021_neurips_albef,2016_CVPR_generation} to evaluate on the RefCOCO+ dataset~\cite{2016_eccv_refcocoplus} in a weakly-supervised setting. 

Table~\ref{tab:downstream} reports the performance comparisons of different vision-language pre-training baselines on downstream VQA, VR and VE tasks.
Similar to image-text retrieval task, MixGen consistently boosts performance on the three tasks.
Under the 3M setting, ALBEF with MixGen outperforms its corresponding baseline by absolute $0.28\%$ on VQA test-std, $0.89\%$ on NLVR$^{2}$ test-P, and $0.36\%$ on SNLI-VE test.
Note that ALBEF with MixGen under 3M setting even outperforms original ALBEF trained with 4M images on VQA and NLVR$^{2}$.

\begin{table*}[t]
\begin{center}
\caption{\textbf{MixGen variants.} For (d) and (e), $|T|_{\lambda}$ indicates randomly keeping $\lambda \cdot |T|$ tokens in text sequence $T$, where $|T|$ is the number of tokens in this sequence. See text for more details.}
\label{tab:mixgen_variants}
\vspace{1ex}
\scalebox{0.9}{
\begin{tabular}{c|c|c|c}  
\toprule
Variants & Image & Text & $\lambda$ \\
\midrule
MixGen & $I_{k} = \lambda \cdot I_{i} + (1 - \lambda) \cdot I_{j}$ & $T_{k} = \mathrm{concat}(T_{i}, T_{j})$ & $\lambda=0.5$ \\
(a) & $I_{k} = \lambda \cdot I_{i} + (1 - \lambda) \cdot I_{j}$ & $T_{k} = \mathrm{concat}(T_{i}, T_{j})$ & $\lambda \sim$ Beta(0.1, 0.1) \\
(b) & $I_{k} = \lambda \cdot I_{i} + (1 - \lambda) \cdot I_{j}$ & $T_{k} = T_{i} \text{ or } T_{j}$ & $\lambda=0.5$ \\
(c) & $I_{k} = I_{i} \text{ or } I_{j}$ & $T_{k} = \mathrm{concat}(T_{i}, T_{j})$ & - \\
(d) & $I_{k} = \lambda \cdot I_{i} + (1 - \lambda) \cdot I_{j}$ & $T_{k} = \mathrm{concat}(|T_{i}|_{\lambda}, |T_{j}|_{1-\lambda})$ & $\lambda \sim$ Beta(0.1, 0.1) \\
(e) & $I_{k} = \lambda \cdot I_{i} + (1 - \lambda) \cdot I_{j}$ & $T_{k} = |\mathrm{concat}(T_{i}, T_{j})|_{0.5}$ & $\lambda \sim$ Beta(0.1, 0.1) \\
\bottomrule
\end{tabular}}
\end{center}
\end{table*}

\begin{table*}
\begin{center}
\caption{\textbf{Ablation of MixGen design}. For image-text retrieval task, we report the RSUM metric. Ft: fine-tuned setting. Zs: zero-shot setting.\label{table:variants}}
\scalebox{0.8}{
  \begin{tabular}{p{2.4cm}|p{2.0cm}p{2.0cm}p{2.0cm}p{1.5cm}p{1.5cm}p{1.5cm}}
  \toprule
  \multirow{2}{*}{MixGen variants}&\makecell[c]{Flickr30K-Ft}&\makecell[c]{MSCOCO-Ft}&\makecell[c]{Flickr30K-Zs}&\makecell[c]{SNLI-VE}&\makecell[c]{NLVR$^2$}&\makecell[c]{VQA} \\
    &\makecell[c]{(1K test set)}&\makecell[c]{(5K test set)}&\makecell[c]{(1K test set)}&\makecell[c]{(test)}&\makecell[c]{(test-P)}&\makecell[c]{(test-dev)}\\
  \hline
  \color{gray}\makecell[c]{ALBEF-base}&\color{gray}\makecell[c]{{555.7}}&\color{gray}\makecell[c]{{470.7}}&\color{gray}\makecell[c]{{518.0}}&\color{gray}\makecell[c]{78.91}&\color{gray}\makecell[c]{78.09}&\color{gray}\makecell[c]{73.62}\\
  \midrule
  \makecell[c]{MixGen}&\makecell[c]{{\textbf{561.0}}}&\makecell[c]{\textbf{477.7}}&\makecell[c]{{\textbf{524.3}}}&\makecell[c]{\textbf{79.65}}&\makecell[c]{\textbf{79.42}}&\makecell[c]{73.84}\\
  \makecell[c]{(a)}&\makecell[c]{{553.2}}&\makecell[c]{{467.4}}&\makecell[c]{{512.9}}&\makecell[c]{78.68}&\makecell[c]{77.22}&\makecell[c]{73.15}\\
  \makecell[c]{(b)}&\makecell[c]{{557.2}}&\makecell[c]{470.0}&\makecell[c]{{516.2}}&\makecell[c]{78.78}&\makecell[c]{78.23}&\makecell[c]{73.76}\\
  \makecell[c]{(c)}&\makecell[c]{555.4}&\makecell[c]{472.3}&\makecell[c]{{518.2}}&\makecell[c]{78.87}&\makecell[c]{78.60}&\makecell[c]{73.34}\\
  \makecell[c]{(d)}&\makecell[c]{555.2}&\makecell[c]{463.7}&\makecell[c]{{506.8}}&\makecell[c]{79.07}&\makecell[c]{78.30}&\makecell[c]{73.29}\\
  \makecell[c]{(e)}&\makecell[c]{{559.2}}&\makecell[c]{477.1}&\makecell[c]{{523.7}}&\makecell[c]{79.36}&\makecell[c]{78.33}&\makecell[c]{\textbf{73.95}}\\
  \bottomrule
  \end{tabular}}
\end{center}
\vspace{-2ex}
\end{table*}

Table~\ref{tab:grounding} reports the performance of visual grounding on RefCOCO+ dataset.
ALBEF$_{itc}$ and ALBEF$_{itm}$ are two variants that compute Grad-CAM visualization for grounding. 
Since ALBEF$_{itm}$ performs better, we adopt ALBEF$_{itm}$ as our baseline and add MixGen on top of it.
We can see that our model with MixGen trained under 3M setting not only surpasses its corresponding baseline on both test sets, but also outperforms ALBEF$_{itm}$ trained with 14M dataset on TestB. 
All empirical results above suggest the superior data efficiency brought by MixGen.

\subsection{Visualizations of MixGen}
\label{subsec:visualization}

\textbf{Image-text retrieval} In Figure~\ref{fig:vis_retrieval}, we show visualization of text-to-image retrieval on MSCOCO.
To be specific, given a text query, we want to compare the rank of the retrieved ground-truth image out of all the retrieved images between ALBEF with and without MixGen. 
We can see that MixGen is usually able to locate the matching image in top-3 retrievals, performing significantly better than the baseline ALBEF.

\textbf{Visual grounding and VQA} In Figure~\ref{fig:vis_vg_vqa}, we show Grad-CAM visualizations to help us understand why MixGen is benficial. 
For the visual grounding task on the RefCOCO+ dataset, we can see that model trained with MixGen can locate image regions more precisely according to the text query. 
Even for the failure case, model trained on MixGen is able to attend better on ``sitting men'', instead of focusing on the wall (wo MixGen model).
For VQA task on VQA2.0 dataset, MixGen can attend to important cues that lead to the correct answer (e.g., black background for predicting if this is a black-white image). 
For the failure case, telling time might be too challenging, where both ALBEF and ALBEF + MixGen fail.

\section{Ablation Studies}
\label{sec:discussion}

In this section, we perform various ablation studies to support the design choice of MixGen.
Unless otherwise stated, we adopt ALBEF as baseline and use the 1M setting which consists of three datasets (COCO, VG, and SBU) in model pre-training.

\noindent \textbf{Design variants}
In Sec.~\ref{subsec:variants}, we introduce 5 variants to support the design choice of MixGen.
Detailed comparisons among variants can be found in Table~\ref{tab:mixgen_variants}.
Here we use the same training setting for all 5 variants and evaluate them across five downstream datasets. 
As we can see in Table~\ref{table:variants}, our default MixGen achieves the overall best performance, and consistently outperform other variants on different tasks except VQA.

\begin{figure*}[t]
\begin{minipage}[t]{0.48\textwidth}
\centering
\scalebox{0.8}{
  \begin{tabular}{l|l|l}
  \toprule
  \multirow{2}{*}{Methods}&\makecell[c]{Flickr30K-Ft}&\makecell[c]{MSCOCO-Ft} \\
    &\makecell[c]{(1K test set)}&\makecell[c]{(5K test set)}\\
     \midrule
  Baseline&555.7&470.7\\
  + Embedding-level &558.9$_{\textcolor{forestgreen}{+3.2}}$ &471.0$_{\textcolor{forestgreen}{+0.3}}$\\
  + Input-level & $\textbf{561.0}_{\textcolor{forestgreen}{+5.3}}$ & $\textbf{477.7}_{\textcolor{forestgreen}{+7.0}}$ \\
  \bottomrule
  \end{tabular}
  }
\end{minipage}
\quad
\begin{minipage}{0.48\textwidth}
\centering
\includegraphics[width=\textwidth]{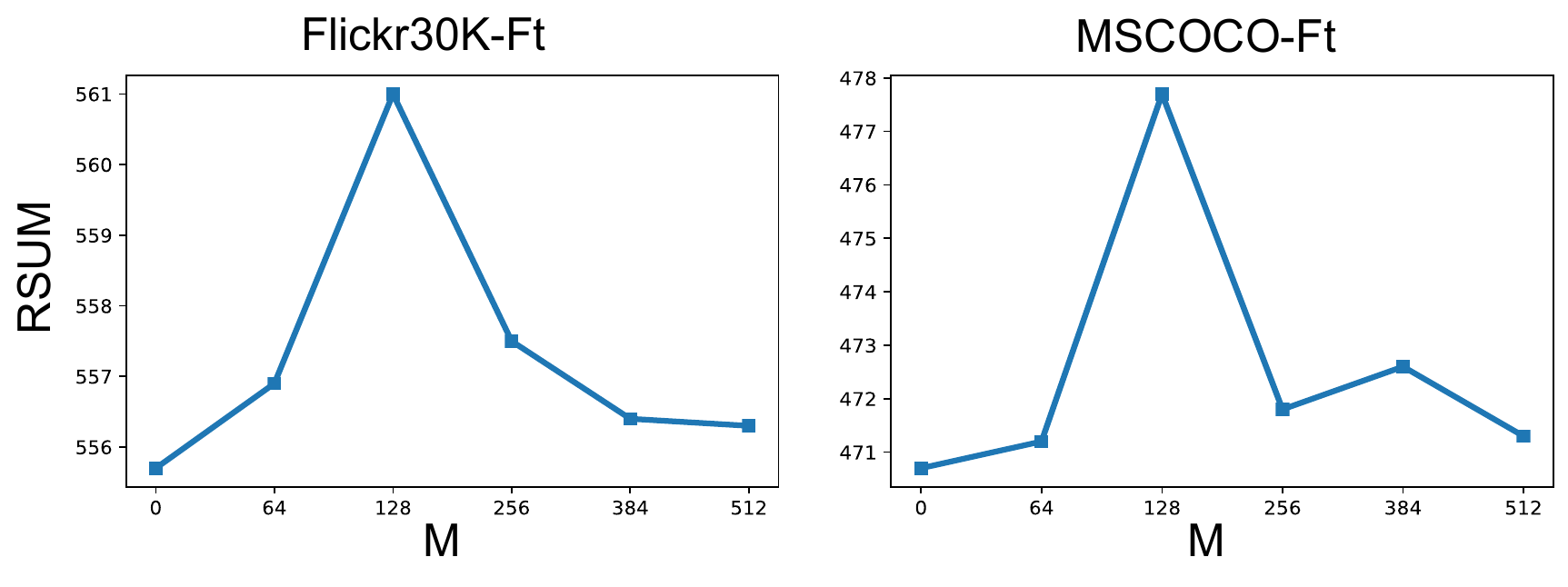}
\end{minipage}
\caption{\textbf{Left:} Input-level vs embedding-level MixGen. Input-level  performs consistently better than embedding-level MixGen. \textbf{Right:} Optimal ratio of MixGen samples. Given batch size $B$=512, we find $M$=128 (i.e., $M$=$B/4$) leads to best performance.\label{fig:optimal_embedding}}
\end{figure*}

\noindent \textbf{Input-level vs embedding-level MixGen}
Recall in Sec.~\ref{subsec:input_embedding} that MixGen can be either performed on the raw input as input-level MixGen, or performed on the embedding level as embedding-level MixGen. 
Here, we report the overall RSUM scores on COCO and Flicker30K datasets.
As we can see in Figure~\ref{fig:optimal_embedding} left, both input-level and embedding-level MixGen perform better than the baseline. 
In addition, input-level MixGen performs consistently better than embedding-level MixGen. 

\noindent \textbf{Optimal ratio of MixGen samples}
In our pseudocode, we replace the first $M$ training samples with the newly generated pairs by MixGen, so that the total batch size and training iterations stay the same. 
Motivated by mixup~\cite{2018_iclr_mixup}, we could randomly shuffle existing pairs and replace them all by new samples, i.e., $M=B$.
Here, we perform an ablation study to investigate how many new pairs is optimal.
As we can see in Figure~\ref{fig:optimal_embedding} right, given a fixed batch size $B=512$ while varying $M$, $M=128$ (i.e., $M = B/4$) leads to the best performance.



\section{Discussion}
\label{sec:limitation}
While we have shown MixGen's effectiveness across multiple settings, there are many more factors that we do not consider in this paper. 
First, there are many popular unimodal data augmentation techniques we could compare to~\cite{pmlr-v119-kim20b,2021_iclr_saliencymix}. 
However, due to its unlimited combination, we can only choose the most related configurations for evaluation, and leave the rest comparisons for future work. 
Second, due to the computational limits, we only perform pre-training on the most widely adopted 4M set and compare to other work on this setting for fair comaprison.
However, there are larger-scale datasets becoming public available recently that we could explore, such as LAION-400M~\cite{laion400M}. 
Third, despite we have evaluated MixGen on four models and five vision-language tasks, there are more models and tasks in vision-language domain. For example, several recent work~\cite{li2022blip,Yu2022CoCaCC,2022_iclr_simvlm} have adopted encoder-decoder architecture and reported performance on generation tasks like image captioning. 
There are also work in video-text retrieval~\cite{Luo2021CLIP4Clip}, text-to-image generation~\cite{dalle2}, language guided object detection~\cite{li2021grounded} or segmentation~\cite{xu2022groupvit}. 
It would be interesting to investigate the generalization capability of MixGen for these methods and tasks.

\section{Related Work}
\label{sec:related_work}

\noindent \textbf{Data augmentation in CV and NLP}
Data augmentation serves an integral part to the success of training most deep networks in CV and NLP, especially when the training set is small.
For CV, the techniques evolve with the model development in the past decade, starting from basic ones (random translation, horizontal flipping, random rotation, color jittering, random resized crop, etc.), to more advanced ones (Cutout~\cite{2017_arxiv_cutout}, mixup~\cite{2018_iclr_mixup}, CutMix~\cite{2019_iccv_cutmix}, random erasing~\cite{2020_aaai_randomerasing}, etc.).
With the rise of AutoML, researchers start to search data augmentation policies automatically from the data, such as AutoAugment~\cite{2019_cvpr_autoaugment} and RandAugment~\cite{2020_neurips_randaugment}. 
For NLP, besides paraphrasing-based methods like back-translation~\cite{xie2020unsupervised}, tokens in text can be replaced by applying thesaurus derived from WordNet~\cite{zhang2015character}, word and framework embeddings~\cite{wang2015s}, and masked language models~\cite{jiao2019tinybert} without changing the meaning of the original text.
Swapping, removal, insertion, substitution are common random operators to add noise to text at both the word level~\cite{wei2019eda,coulombe2018text,regina2020text} and the sentence level~\cite{yan2019data}. 
All in all, data augmentation leads to improved performance, higher data efficiency, and robustness to domain shift and adversarial attack, which is worthy of investigating in the multi-modal domain. 

\noindent \textbf{Vision-language pre-training}
Despite the rapid progress of data augmentation in each uni-modal domain (CV, NLP, speech, etc.), joint multi-modal data augmentation is rarely studied.
Most of the focus from the community has been on modeling improvements~\cite{2020_iclr_vlbert,2021_arxiv_clip,2021_icml_vilt,cho2021vlt5,tan2019lxmert,Kamath2021MDETRM,xu_acl21_e2evlp}, multi-modal fusion techniques~\cite{vilbert_Lu2019ViLBERTPT,2021_arxiv_vlmo,Appalaraju_2021_ICCV,2021_neurips_albef,2020_neurips_villa} and multi-modal loss functions~\cite{2021_arxiv_clip,2021_icml_vilt,TCL_Yang2022VisionLanguagePW,ipot_pmlr-v115-xie20b,zolfaghari_iccv2021_crossclr}.
Existing works often perform mild data augmentation on either vision or text modality, but not jointly.
There are a few related work on joint augmentation. \cite{2020_eccv_seadavqa} has proposed to generate semantically equivalent adversarial examples of both visual and textual data as augmented training samples for VQA. 
\cite{2021_intelligent_mda} learns a cross-modality matching network to select image–text pairs from existing uni-modal datasets as the multi-modal synthetic dataset, and uses this dataset to enhance the performance of classifiers for sentiment analysis. 
Both approaches are effective, but only designed for specific vision-language downstream tasks.
On the contrary, our proposed MixGen is simple, effective, and compatible with generic vision-language pre-training methods.

\noindent \textbf{Relation to mixup}
Mixup was originally introduced as a data-agnostic augmentation technique for CV~\cite{2018_iclr_mixup}.
It is a widely adopted (even by-default) technique for training CNNs or vision transformers. 
Specifically, mixup generates new training samples via
\begin{equation}
\begin{split}
    x_{k} &= \lambda \cdot x_{i} + (1 - \lambda) \cdot x_{j} \\ 
    y_{k} &= \lambda \cdot y_{i} + (1 - \lambda) \cdot y_{j},
\end{split}
\end{equation}
where $x_{i}$ and $x_{j}$ are raw input vectors, $y_{i}$ and $y_{j}$ are one-hot label encodings, and $\lambda \in [0, 1]$ is sampled from a Beta distribution. 
Recently, \cite{2019_arxiv_textmixup} adapted mixup to the NLP domain by replacing raw image pixels $x$ with text embeddings. 

MixGen differs from mixup in several aspects.
First, MixGen is proposed for multi-modal data augmentation. 
The input to MixGen is data from different modalities, and there is no one-hot label encodings $y$ involved in Equation~(\ref{equ:input_mixgen}). 
Second, MixGen does not perform mixup on the text modality. 
Instead, MixGen simply concatenates text sequences to perform image-text co-augmentation and best preserves information.

\section{Conclusion}
\label{sec:conclusion}

In this work, we present a new vision-language joint data augmentation method termed MixGen.
Adding MixGen on four recent state-of-the-art models achieves consistent improvement across five different downstream tasks.
Strong empirical results suggest that MixGen not only makes these models learn better multi-modal latent representations, but also improves their data efficiency.
We hope that this work will provide useful data points for future research on joint multi-modal data augmentation.

{\small
\bibliographystyle{ieee_fullname}
\bibliography{egbib}
}

\end{document}